\documentclass{article}
\usepackage{iclr2027_conference,times}
\usepackage[hyphens]{url}
\usepackage{graphicx}
\usepackage[style=base]{caption}
\usepackage{anyfontsize}
\usepackage{amsmath}
\usepackage{amssymb}
\usepackage{booktabs}
\usepackage{array}
\usepackage{tabularx}
\usepackage{placeins}
\usepackage{xcolor}
\usepackage{hyperref}
\hypersetup{hidelinks,pdfauthor={Bo Deng, Kang Zhou, Lifan Guo, Chongyang Tao, Xuanren Chen, Chenggang Xie, Renzhao Liang, Feng Chen, Chi Zhang}}
\newcommand{\evolve}{\mathrm{evo}}
\newcommand{\noevolve}{\mathrm{noevo}}

\title{FinEvo-Bench: A Longitudinal Benchmark\\
for Self-Evolving Agents in\\
Professional Financial Workflows}
\author{
Bo Deng\textsuperscript{1,2}\thanks{This work was conducted during an internship at Alibaba Cloud Computing.},
Kang Zhou\textsuperscript{2},
Lifan Guo\textsuperscript{2},
Chongyang Tao\textsuperscript{1}\thanks{Corresponding author.},\\
Xuanren Chen\textsuperscript{1},
Chenggang Xie\textsuperscript{1},
Renzhao Liang\textsuperscript{1},
Feng Chen\textsuperscript{2},
Chi Zhang\textsuperscript{2}\\
\textsuperscript{1}Beihang University
\qquad
\textsuperscript{2}Qwen DianJin Team, Alibaba Cloud Computing
}
\iclrfinalcopy

\begin{document}
\maketitle
\lhead{}

\begin{abstract}
Agents used over time encounter recurring professional work: each case requires different evidence and judgment, while the underlying workflow can be reused. Benchmarks built from independent tasks cannot reveal whether an agent turns earlier experience into better procedures for later cases. We introduce FinEvo-Bench, a longitudinal benchmark designed around this structure. It contains 120 open-ended tasks drawn from real cases across 20 business scenes in six financial domains. Each scene contains six substantively different cases that share a professional workflow and an expert-authored rubric for task quality and financial compliance. Constructing and validating the benchmark required approximately 1,200 person-hours. Finance provides a natural test bed because recurring analyses apply shared professional and compliance requirements to heterogeneous inputs, producing case-specific analyses and conclusions. We evaluate four self-evolving agent scaffolds with Qwen3.7-Max on three independently shuffled, globally interleaved task streams. A Claude Code rubric judge backed by Claude Opus~4.6 evaluates all outputs, and paired state-reset controls estimate each scaffold's gain from retained experience. Evolving runs score 9.33--19.37 points higher and trigger 0.12--0.44 fewer compliance issues per task than their paired controls. Paired score gains at within-scene ranks~4--6 exceed those at ranks~1--3 by 6.10--8.70 points. FinEvo-Bench measures whether retained experience improves later professional work under continued use.

\end{abstract}

\section{Introduction}

Agents deployed in professional settings are used repeatedly. A credit analyst reviews one borrower after another; a claims specialist applies the same adjudication process to new claim files. The cases differ in their inputs, required analyses, and valid conclusions, but professional procedures recur. An agent improves under continued use when it turns earlier experience into a better procedure for the next case. Self-evolving agents pursue this behavior through persistent memories, skills, or other state \citep{shinn2023reflexion,zhao2024expel,wang2023voyager,zhang2026ace}.

Most agent benchmarks evaluate tasks independently and therefore observe current execution capability rather than growth across successive tasks. Recent self-evolution benchmarks introduce task streams, persistent environments, or iterative artifact optimization \citep{zheng2025lifelongagentbench,jiang2026seaeval,wei2025evomemory,cai2025stulife,chi2026frontiereng}. Yet evaluating continued professional use requires a specific task structure: later tasks must preserve enough procedural continuity for experience to transfer while differing in their case-specific evidence and valid conclusions. Their outputs must also remain open-ended, because professional reports, assessments, and recommendations rarely have a single valid realization.

\begin{table}[!ht]
\centering
\caption{Representative self-evolution benchmarks.}
\label{tab:self_evolution_benchmarks}
{\small
\setlength{\tabcolsep}{2.1pt}
\renewcommand{\arraystretch}{0.98}
\begin{tabular}{@{}>{\raggedright\arraybackslash}p{0.29\columnwidth}cccc@{}}
\toprule
Benchmark & \shortstack{Cross-\\task\\Evolution} & \shortstack{Domain\\Workflow} & \shortstack{Open-\\ended\\Artifact} & \shortstack{Multi-\\aspect\\Evaluation} \\
\midrule
LifelongAgent\allowbreak{}Bench & \checkmark & & & \\
SEA-Eval & \checkmark & & & \checkmark \\
Evo-Memory & \checkmark & & & \checkmark \\
StuLife & \checkmark & \checkmark & & \checkmark \\
Frontier-Eng & & \checkmark & \checkmark & \checkmark \\
\midrule
\textbf{FinEvo-Bench (Ours)} & \checkmark & \checkmark & \checkmark & \checkmark \\
\bottomrule
\end{tabular}
}

\end{table}

Table~\ref{tab:self_evolution_benchmarks} compares representative benchmarks; Section~2 covers the broader literature. FinEvo-Bench targets recurring professional workflows with open-ended artifacts and multidimensional evaluation. Finance is particularly suitable for this setting. Credit review, claim analysis, insurance advice, and investment research apply stable analytical and compliance procedures to cases whose files, analytical demands, and appropriate conclusions differ sharply. This creates a concrete test of whether an agent can reuse a professional process across substantively different cases. Existing financial benchmarks evaluate rich professional artifacts \citep{jin2026finrpt,kundurthy2026bluefin}, but score cases independently.

FinEvo-Bench contains 120 multi-file tasks grounded in institutional or publicly documented real cases, organized as six cases for each of 20 scenes across six financial domains. Domain experts construct and cross-review the tasks, reference answers, and scene-level rubrics; case curation, task construction, and validation required approximately 1,200 person-hours. The automated rubric judge achieves high absolute agreement with a financial expert ($\mathrm{ICC(A,1)}=0.95$; 95\% CI: $[0.93,0.97]$).

To model continued use, we independently shuffle the tasks into three globally interleaved streams. Related cases recur after unrelated intervening work, requiring each scaffold to retain and retrieve the relevant experience. We use the same model and inference configuration for every scaffold and pair each evolving run with a state-reset control. The protocol therefore reports final task performance and the improvement associated with retained experience separately.

Across four scaffolds, evolving runs score 9.33--19.37 points higher and trigger 0.12--0.44 fewer compliance issues per task than their state-reset controls. For every scaffold, mean gains over within-scene ranks~4--6 exceed those over ranks~1--3 by 6.10--8.70 points.

Our contributions are:
\begin{itemize}
    \item FinEvo-Bench provides 120 open-ended, multi-file tasks grounded in real financial cases and validated through expert construction and review.
    \item Each business scene combines a shared professional workflow with six substantively different cases and an expert-authored rubric for task quality and financial compliance.
    \item A paired, interleaved protocol measures whether retained experience improves later work, and evaluation across four agent scaffolds reports final capability, self-evolution gain, experience use, and cost.
\end{itemize}

\section{Related Work}

\paragraph{Self-evolution.}
Self-evolving agents use task outcomes to improve later behavior. They may retain reflections or memories \citep{shinn2023reflexion,zhao2024expel}, compile reusable skills or playbooks \citep{wang2023voyager,zhang2026ace}, or adapt workflows and model policies \citep{zhang2024agentoptimizer,zhang2025aflow,wang2025ragen}. FinEvo-Bench accommodates these different persistent representations and uses paired non-evolving controls to isolate gains attributable to retained experience.

\paragraph{Longitudinal evaluation of evolving agents.}
Conventional web, computer, and software-engineering benchmarks evaluate tasks independently and do not preserve experience from prior tasks across episodes \citep{zhou2024webarena,xie2024osworld,jimenez2024swebench}. Recent benchmarks instead expose task sequences or persistent memory so that adaptation beyond isolated episodes can be assessed. LifelongAgentBench constructs skill-dependent interactive tasks \citep{zheng2025lifelongagentbench}, while SEA-Eval evaluates correlated and orthogonal streams through success and token trajectories \citep{jiang2026seaeval}. Evo-Memory and EvoMemBench organize their protocols around memory retrieval, update, and reuse; SEAGym separates update evidence from held-out transfer and replay assessment \citep{wei2025evomemory,wang2026evomembench,zheng2026seagym}. FinEvo-Bench combines recurring financial workflows based on real cases, open-ended professional deliverables, and expert-derived rubric evaluation within a globally interleaved stream.

\paragraph{Financial benchmarks.}
Financial benchmarks cover numerical and document reasoning \citep{chen2021finqa,zhu2021tatqa,chen2022convfinqa,islam2023financebench}, broad capability suites \citep{xie2023pixiu,xie2024finben}, and professional analysis, research, tool use, or spreadsheet tasks \citep{benyoash2025secque,bigeard2025financeagent,choi2025finagentbench,jin2026finrpt,zhu2026finmcp,kundurthy2026bluefin}. FinRpt's multidimensional evaluation and BlueFin's granular criteria accommodate professional artifacts with multiple valid forms, but their instances are still scored independently. FinEvo-Bench interleaves related but distinct cases, isolates gains from retained experience, and separately reports quality and compliance.

\FloatBarrier

\section{FinEvo-Bench}

\subsection{Benchmark Scope and Design}

\begin{figure}[!t]
  \centering
  \includegraphics[width=0.95\textwidth]{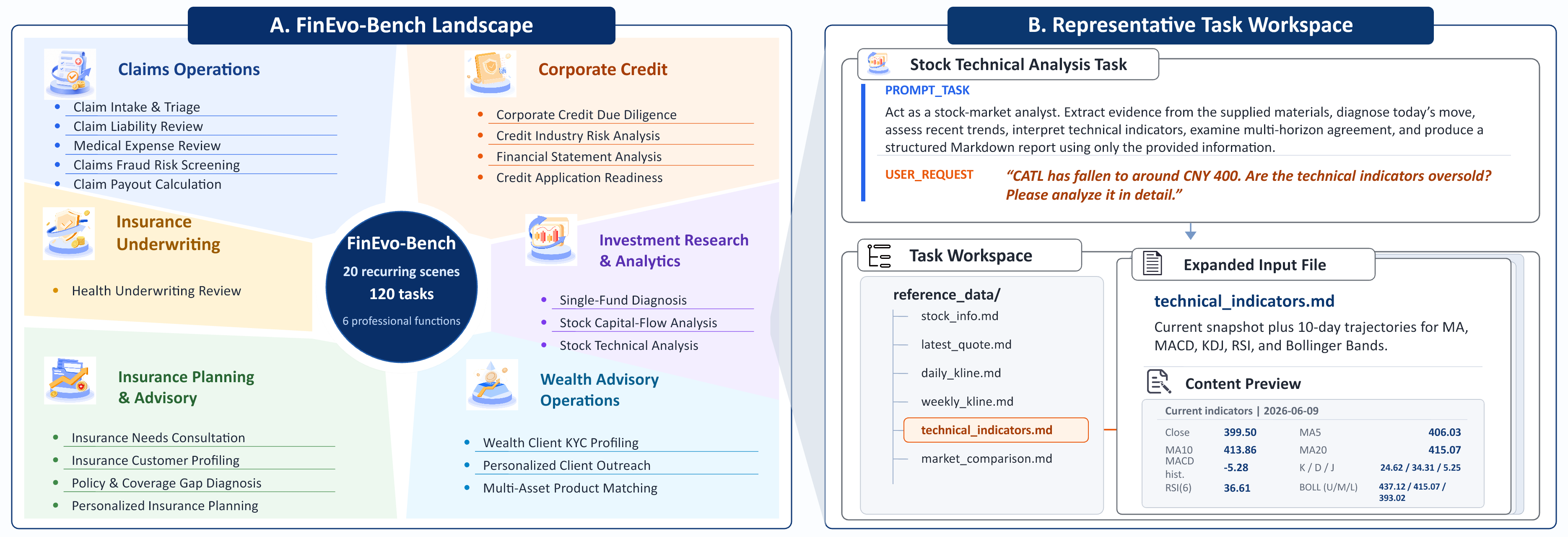}
  \caption{Overview of FinEvo-Bench. Left: 20 recurring financial workflows across six domains, each instantiated as six real-case tasks. Right: a representative open-ended Stock Technical Analysis task with its user request and six input files.}
  \label{fig:dataset_overview}
\end{figure}

FinEvo-Bench organizes longitudinal evaluation around the \emph{business scene}, a professional workflow that recurs across different cases. It contains 20 scenes across the six financial domains shown in Figure~\ref{fig:dataset_overview}.

Each scene contains six substantively distinct cases governed by the same professional procedure. The shared procedure creates a basis for transferring experience, while different files, evidence patterns, analytical demands, and conclusions require fresh case-level analysis. The 120 tasks contain 775 input files, averaging 6.46 per task (range: 2--11), and request open-ended reports, assessments, or recommendations. Scene-specific rubrics cover task-specific professional requirements, including information and evidence use, analysis and calculation, conclusions and recommendations, report quality, and financial compliance, while accepting multiple valid outputs.

\subsection{Scene Definition and Task Construction}

We construct each scene $s$ from three sources: an institutional scene description $D_s$, a reference professional procedure $P_s$ validated in practice, and a candidate pool $\mathcal{C}_s$ of institutional and publicly documented real cases. The scene description defines the business problem. The reference procedure specifies the required input files, main steps, expected deliverable, and professional constraints. Individual cases supply the facts, data, and judgment conditions. We review source permissions and release conditions before admitting a case to the eligible pool $\mathcal{C}^{\mathrm{eligible}}_s$. When necessary, we remove or replace direct identifiers and sensitive fields while preserving data relationships, numerical logic, chronology, and decision conditions.

A domain expert consolidates these sources into a scene specification:
\begin{equation}
\mathcal{S}_s=\Phi(D_s,P_s,\mathcal{C}^{\mathrm{eligible}}_s)
            =(O_s,E_s,A_s,Y_s,K_s).
\label{eq:scene_specification}
\end{equation}
Here, $O_s$ is the business objective and scope; $E_s$, the required inputs; $A_s$, the professional operations and checks; $Y_s$, the expected deliverable; and $K_s$, the compliance requirements and professional boundaries.

After defining a scene, experts select six cases from its eligible pool. The selection follows the scene's business characteristics and professional procedure. Cases must differ substantively in their business situations, input files, analytical focus, judgment conditions, or conclusions; cases with only superficial differences are excluded from the same scene. Table~\ref{tab:scene_task_examples} gives four examples.

\begin{table}[t]
\centering
\caption{Six-case designs for selected scenes.}
\label{tab:scene_task_examples}
{\small
\setlength{\tabcolsep}{3pt}
\renewcommand{\arraystretch}{0.98}
\begin{tabularx}{\columnwidth}{@{}>{\raggedright\arraybackslash}p{0.25\columnwidth} >{\raggedright\arraybackslash}X@{}}
\toprule
Scene & Representative differences across the cases \\
\midrule
Financial Statement Analysis & Healthy operations; cyclical downturn; divergence between earnings and cash flow; reporting red flags; multiple distress signals \\
Claim Payout Calculation & Standard calculation; substantial expense disallowance; ineligible hospitalization; deductible threshold; data anomaly; repeated claims reaching the coverage limit \\
Single-Fund Diagnosis & High-performing equity fund; persistently underperforming fund; index fund; pure bond fund; mixed bond fund \\
Personalized Client Outreach & Client-information update; risk reminder; product maturity and rollover; event invitation; routine review; outreach with incomplete client information \\
\bottomrule
\end{tabularx}
}

\end{table}

Each selected case becomes a task with a natural-language request, input files, task metadata, and a reference answer. Domain experts review the final tasks in two stages. First, they assess the scene and task settings. They then verify that the input files are complete and reflect actual business conditions, and that the reference answer has correct calculations, analysis, and conclusions. Appendix A gives a construction example with task files, reference answer, rubric, and review.

The worked Financial Statement Analysis scene in Appendix~A illustrates the resulting variation. Its six cases range from a healthy manufacturer seeking working capital to a distressed trader and a private company with incomplete reporting. A medium--hard case presents the agent with a request and ten files covering statements, notes, audit findings, related parties, non-recurring items, and industry comparisons. The cases share the same analytical procedure, but require different evidence checks, financial analyses, and credit recommendations.

\subsection{Scene-Level Rubrics and Quality Control}

The six tasks in a scene share a 100-point rubric derived from the scene's professional procedure. The rubric checks whether an agent uses the input files correctly, completes the necessary analysis, and reaches supported conclusions while accepting multiple valid structures and formulations.

The rubric also checks financial compliance, including fabricated or unsupported data and terms, definitive claims based on insufficient information, decisions beyond the agent's professional role, and guarantees about credit approval, claim outcomes, or investment returns. Two domain experts construct each rubric. One drafts the criteria, point allocations, and grading rules. The other applies the rubric to all six tasks to check case applicability, coverage of required business steps, and consistency with the input files and reference answers. They resolve any omission or conflict and fix the rubric before the experimental runs. Section~\ref{sec:evaluator_validation} separately validates the automated application of these rubrics against a financial expert's scores.

The worked rubric in Appendix~A contains 27 task criteria and five report-quality criteria. Its rules define full, partial, and zero credit, applicability conditions, numerical tolerances, evidence requirements, and critical failures. During evaluation, the judge receives the case files, final deliverable, and rubric. The rubric awards credit for correctly completing the required analyses and reaching conclusions supported by the case evidence, while accepting valid variations in wording and structure.

Across all 20 scenes, case curation, de-identification, input-file assembly, reference-answer development, rubric construction, and expert cross-review required approximately 1,200 person-hours.

\FloatBarrier

\section{Experiments}

\subsection{Experimental Setup}

\begin{figure}[t]
  \centering
  \includegraphics[width=0.95\textwidth]{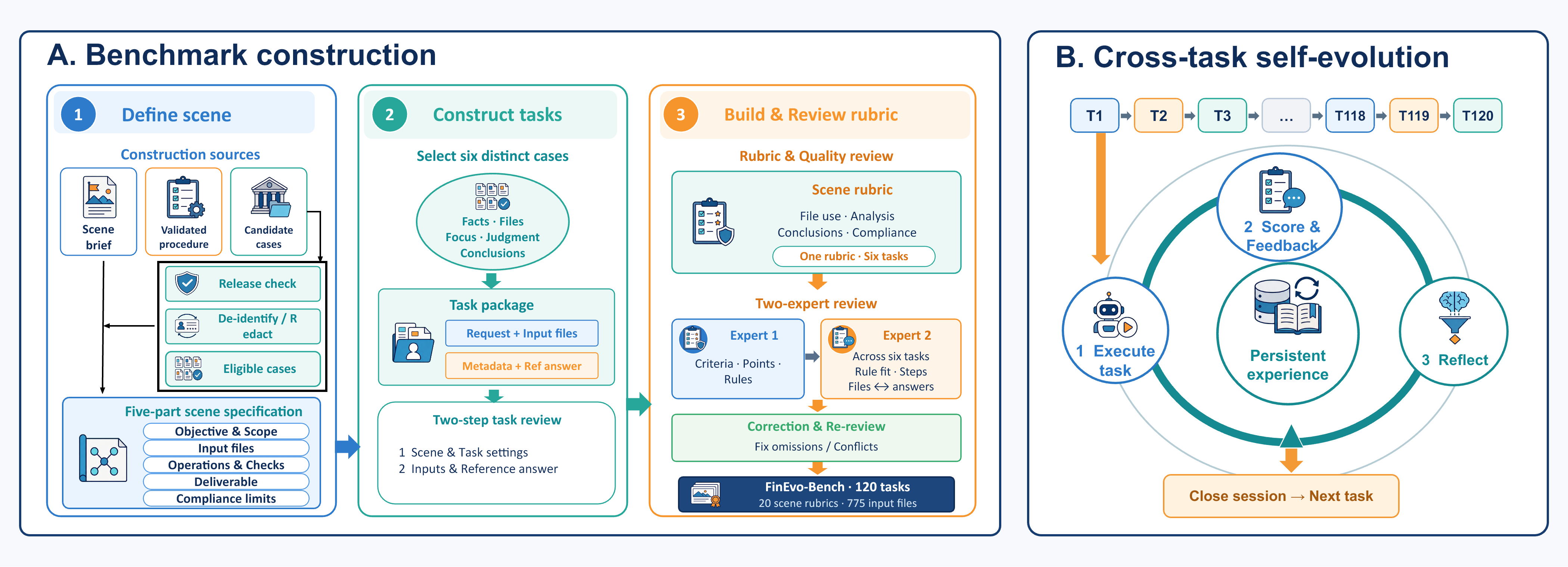}
  \caption{FinEvo-Bench construction (A) and experimental setup for cross-task self-evolution (B).}
  \label{fig:construction_protocol}
\end{figure}

We evaluate four self-evolving agent scaffolds with matched evolving and state-reset runs under the longitudinal protocol in Figure~\ref{fig:construction_protocol}.

\paragraph{Self-Evolving Agent Scaffolds.}
The four scaffolds retain experience in different forms. Claude Code and Codex can distill task experience into reusable skills and project-scoped memory. Letta maintains editable memory blocks for each agent. Its memory-management tools can insert, replace, or rewrite content, and every model call receives the full contents of all core memory blocks. GenericAgent distills each completed task into a reusable Markdown experience file. A prompt-resident title-only L0 index guides selective loading of full files \citep{liang2026genericagent}.

\paragraph{Longitudinal Evaluation Protocol.}
We independently shuffle the 120 tasks three times to form three globally interleaved streams. Each evolving run starts without benchmark experience and processes tasks sequentially in stream order. Each task cycles through execution, scoring and feedback, reflection, and consolidation \citep{shinn2023reflexion,zhao2024expel,zhang2026ace}. After the scaffold submits its deliverable, a separate Claude Code scoring agent backed by Claude Opus~4.6 scores it using the scene rubric and returns targeted diagnostic feedback for reflection and experience consolidation. We close the session, removing raw conversation history while preserving the updated experience for subsequent tasks.

The paired non-evolving condition resets agent state before every task. Within each run, both conditions share the task order, backbone, decoding configuration, and scoring procedure; only retained experience differs.

\paragraph{Feedback and Experience Updates.}
The scoring agent applies the expert-authored scene rubric and converts its record into concise diagnostic feedback identifying the specific deficiencies responsible for lost credit. During reflection, the scaffold converts these task-specific diagnostics into persistent memories or skills for subsequent tasks. Appendix~\ref{app:complete_example} traces this process from deliverable to experience update.

\paragraph{Evaluation Metrics.}
We report four metrics. (1) \emph{Task quality} is the mean scene-specific rubric score on a 0--100 scale. (2) \emph{Financial compliance} is the mean number of triggered compliance issues per task. (3) \emph{Self-evolution ability} is measured by the paired score gain and reduction in compliance issues relative to the non-evolving condition. Let $S_{a,k,i}^{\evolve}$ and $S_{a,k,i}^{\noevolve}$ denote the scores of scaffold $a$ in run $k$ on task $i$, and let $C_{a,k,i}^{\evolve}$ and $C_{a,k,i}^{\noevolve}$ denote the corresponding numbers of compliance issues. For $K=3$ runs and $N=120$ tasks,
\noindent
\begin{minipage}{0.50\linewidth}
\begin{equation}
\Delta\mathrm{Score}_a
=\frac{1}{KN}\sum_{k=1}^{K}\sum_{i=1}^{N}
\left(S_{a,k,i}^{\evolve}-S_{a,k,i}^{\noevolve}\right).
\end{equation}
\end{minipage}%
\begin{minipage}{0.50\linewidth}
\begin{equation}
\Delta\mathrm{Comp.}_a
=\frac{1}{KN}\sum_{k=1}^{K}\sum_{i=1}^{N}
\left(C_{a,k,i}^{\noevolve}-C_{a,k,i}^{\evolve}\right).
\end{equation}
\end{minipage}
(4) \emph{Agent-side cost} is the mean number of tokens consumed during task execution and post-evaluation reflection. We report token counts in units of $10^4$ per task, with total agent-side tokens calculated as the sum of execution and reflection tokens.

For longitudinal evolution, we rank each scene's six tasks by their order in the global stream, so rank $r$ follows $r-1$ tasks from that scene. With $K=3$ runs, $M=20$ scenes, and score $S_{a,k,s,r}$, the mean paired gain is
\begin{equation}
\Delta\mathrm{Score}_{a,r}
=\frac{1}{KM}\sum_{k=1}^{K}\sum_{s=1}^{M}
\left(
S_{a,k,s,r}^{\evolve}
-S_{a,k,s,r}^{\noevolve}
\right).
\end{equation}
We define $\Delta\mathrm{Comp.}_{a,r}$ analogously, using the difference in compliance issues $C_{a,k,s,r}^{\noevolve}-C_{a,k,s,r}^{\evolve}$. Table~\ref{tab:dynamics} summarizes ranks~1--3 as Early and ranks~4--6 as Late, with score gains
\noindent
\begin{minipage}{0.50\linewidth}
\begin{equation}
\Delta\mathrm{Score}^{\mathrm{early}}_a
=\frac{1}{3}\sum_{r=1}^{3}\Delta\mathrm{Score}_{a,r}.
\end{equation}
\end{minipage}%
\begin{minipage}{0.50\linewidth}
\begin{equation}
\Delta\mathrm{Score}^{\mathrm{late}}_a
=\frac{1}{3}\sum_{r=4}^{6}\Delta\mathrm{Score}_{a,r}.
\end{equation}
\end{minipage}
Early and Late compliance reductions use the same rank averages of $\Delta\mathrm{Comp.}_{a,r}$. Ranks are induced independently in each run. We mark \emph{experience activation} whenever stored experience enters the execution context. Claude Code, Codex, and GenericAgent retrieve it selectively, whereas Letta injects all core memory blocks automatically. Letta's 120/120 count reflects its always-on interface. We report activation counts and scores for activated and non-activated subsets.

\paragraph{Implementation Details.}
All four scaffolds use Qwen3.7-Max with a 1M-token context window, a maximum output length of 64K tokens, greedy decoding, and temperature zero. The Claude Code rubric judge is backed by Claude Opus~4.6. We run each full-benchmark configuration three times and report the mean. Appendix B gives the three-stage pipeline code and prompts.

\paragraph{Independent Human Validation of Rubric Application.}
\label{sec:evaluator_validation}

The Claude Code rubric judge and a financial expert independently score the 120 deliverables from one complete run of the main experiment using the same expert-authored scene rubrics. The expert scores are collected for evaluator validation and do not enter the agent feedback loop. Under a two-way mixed-effects absolute-agreement model, the judge achieves $\mathrm{ICC(A,1)}=0.95$ (95\% CI: $[0.93,0.97]$). ICC here measures agreement in absolute scores. The judge's mean and maximum absolute differences from expert scores are 1.6 and 5 points, respectively, on the 0--100 scale. On this validation set, the automated scores show close absolute agreement with the financial expert's application of the predefined professional criteria.

\subsection{Main Results}

\paragraph{Overall Performance and Efficiency.}
Table~\ref{tab:main} reports evolved performance and cost, with paired gains over controls.

\FloatBarrier
\begin{table}[t]
\centering
\caption{Performance, self-evolution gain, and agent-side cost by scaffold.}
\label{tab:main}
\small
\setlength{\tabcolsep}{2.8pt}
\begin{tabular}{lrrrrrrr}
\toprule
& \multicolumn{2}{c}{Evolved performance} & \multicolumn{2}{c}{Overall gain} & \multicolumn{3}{c}{Cost ($10^4$ tokens/task)} \\
\cmidrule(lr){2-3}\cmidrule(lr){4-5}\cmidrule(lr){6-8}
Agent scaffold & Score $\uparrow$ & Comp. $\downarrow$ & $\Delta$Score $\uparrow$ & $\Delta$Comp. $\uparrow$ & Exec. $\downarrow$ & Reflect. $\downarrow$ & Total $\downarrow$ \\
\midrule
Claude Code  & 89.47 & 0.11 & +17.89 & 0.44 & 16.31 & 60.19 & 76.50 \\
Codex        & 91.17 & 0.11 & +19.37 & 0.44 & 20.53 & 48.22 & 68.75 \\
Letta        & 91.65 & 0.09 & +17.82 & 0.39 & 32.56 & 17.87 & 50.43 \\
GenericAgent & 83.34 & 0.34 & +9.33 & 0.12 & 11.57 & 10.21 & 21.78 \\
\bottomrule
\end{tabular}
\end{table}

Across the three independently shuffled task streams, all four scaffolds show positive paired gains: evolved scores span 83.34--91.65, score gains span 9.33--19.37 points, compliance issues fall by 0.12--0.44 per task, and agent-side costs span $21.78\times10^4$--$76.50\times10^4$ tokens per task. Letta attains the strongest evolved performance, Codex the largest paired gain, and GenericAgent the lowest cost; the scaffold ranking therefore depends on whether the target is final quality, improvement through experience, or efficiency.

Letta and GenericAgent use fewer reflection tokens than Claude Code and Codex, but their execution costs differ sharply. Letta uses $17.87\times10^4$ reflection tokens and $32.56\times10^4$ execution tokens per task because every core memory block enters each model call. GenericAgent uses only $10.21\times10^4$ and $11.57\times10^4$, respectively, because it writes compact experience files and loads them selectively. Its total cost is less than one third of Claude Code's. In this comparison, the two scaffolds designed around self-evolution consume fewer tokens during experience consolidation. Agent-side cost depends on both reflection and how stored experience enters the execution context.

\paragraph{Longitudinal Evolution.}
\begin{table}[t]
\centering
\caption{Longitudinal gains and experience utilization by agent scaffold.}
\label{tab:dynamics}
\label{tab:activation}
\small
\setlength{\tabcolsep}{2.0pt}
\begin{tabular}{lrrrrrrrr}
\toprule
& \multicolumn{4}{c}{Longitudinal evolution} & \multicolumn{4}{c}{Experience utilization} \\
\cmidrule(lr){2-5}\cmidrule(lr){6-9}
& \multicolumn{2}{c}{Score gain} & \multicolumn{2}{c}{Comp. reduction} & & & & \\
\cmidrule(lr){2-3}\cmidrule(lr){4-5}
Agent scaffold & Early $\uparrow$ & Late $\uparrow$ & Early $\uparrow$ & Late $\uparrow$ & Activated & Act. score & Non-act. & Gap \\
\midrule
Claude Code  & 14.28 & 21.52 & \textbf{0.43} & 0.45 & 87/120  & 91.82 & 83.29 & +8.53 \\
Codex        & \textbf{15.12} & \textbf{23.62} & 0.41 & \textbf{0.47} & 102/120 & 92.42 & 84.11 & +8.31 \\
Letta        & 13.47 & 22.17 & 0.35 & 0.43 & 120/120 & 91.65 & -- & -- \\
GenericAgent & 6.28 & 12.38 & 0.07 & 0.17 & 71/120 & 87.04 & 77.98 & +9.06 \\
\bottomrule
\end{tabular}
\end{table}

\begin{figure}[t]
\centering
\begin{minipage}[t]{0.47\linewidth}
  \vspace{0pt}
  \centering
  \includegraphics[width=\linewidth]{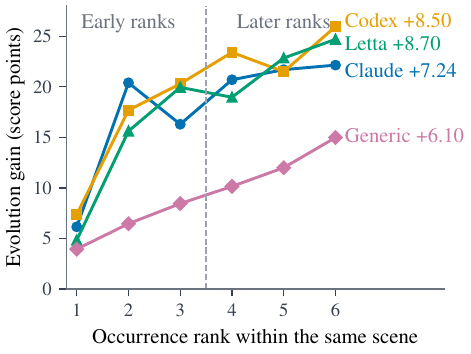}
\end{minipage}\hfill
\begin{minipage}[t]{0.47\linewidth}
  \vspace{0pt}
  \centering
  \includegraphics[width=\linewidth]{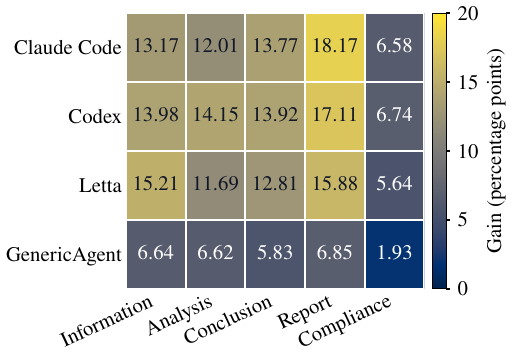}
\end{minipage}
\caption{Self-evolution diagnostics over three runs. Left: score gain by within-scene rank. Right: gains across five normalized dimensions over 120 tasks.}
\label{fig:evolution_diagnostics}
\label{fig:rank_gain}
\label{fig:dimension_gains}
\end{figure}

Figure~\ref{fig:evolution_diagnostics} (left) shows higher gains at later within-scene ranks: every scaffold gains 6.10--8.70 points more over ranks~4--6 than over ranks~1--3, with late gains ranging from 12.38 to 23.62 points. Compliance reductions also increase by 0.02--0.10 issues per task (Table~\ref{tab:dynamics}). As agents process more cases within a scene, their accumulated experience becomes more complete and operationally useful, producing larger quality and compliance gains on later tasks.

The early--late pattern appears in all three independently shuffled task orderings.

\paragraph{Experience Utilization.}
In Table~\ref{tab:dynamics}, Gap is the score for the activated subset minus the score for the subset without activation. Letta loads all core memory blocks on every task. Codex and Claude Code selectively use skills or project-scoped memory on 102 and 87 tasks, respectively. GenericAgent's title-only L0 entries often contain a single keyword, so its lookup fails to surface relevant experience files on many tasks and activates a full file only 71 times.

For the three selective-retrieval scaffolds, activated tasks score 8.31--9.06 points higher than tasks without activation. The subsets without activation score 77.98--84.11, which is 3.97--12.31 points above the corresponding averages under state reset. This advantage suggests that agents sometimes skip retrieval when they are confident that they can complete a task directly. GenericAgent's missed lookups show that non-activation can also result from retrieval failure, making retrieval behavior a key part of scaffold performance.

\subsection{Diagnostic Analyses}

\paragraph{Experience Carriers in Claude Code.}
Claude Code supports both persistent memory and reusable skills. We compare memory only, skills only, and their unrestricted combination over the full benchmark. No evolution and a fixed expert skill derived from each scene's reference workflow serve as references. Costs are reported in $10^4$ tokens per task; reference settings omit reflection.

\begin{table}[t]
\centering
\caption{Experience-carrier configurations in Claude Code.}
\label{tab:carriers}
\small
\setlength{\tabcolsep}{2.8pt}
\begin{tabular}{lrrrr}
\toprule
& & & \multicolumn{2}{c}{Cost ($10^4$ tokens/task)} \\
\cmidrule(lr){4-5}
Setting & Score $\uparrow$ & Comp. $\downarrow$ & Exec. $\downarrow$ & Reflect. $\downarrow$ \\
\midrule
No evolution    & 71.58 & 0.55 & \textbf{14.12} & -- \\
Fixed expert skill & 86.67 & 0.13 & 15.92 & -- \\
Memory + skills & 89.47 & 0.11 & 16.31 & 60.19 \\
Memory only     & 90.42 & 0.09 & 17.94 & \textbf{26.18} \\
Skill only      & \textbf{93.71} & \textbf{0.05} & 17.53 & 44.03 \\
\bottomrule
\end{tabular}

\end{table}

The skill-only configuration reaches 93.71 with 0.05 compliance issues per task (Table~\ref{tab:carriers}). It concentrates evidence checks, analytical steps, report structures, and compliance constraints in a reusable procedure suited to recurring financial work. The memory-only configuration also improves substantially over no evolution, reaching 90.42 with 0.09 issues per task. The unrestricted combination reaches 89.47 with 0.11 issues. The inspected execution traces explain why the combination underperforms skills alone: reflection splits updates across memory and skills, while later tasks often retrieve only one carrier, fragmenting the reusable procedure available during execution. Execution costs remain similar across the three evolving settings, while reflection costs are 26.18 for memory only, 44.03 for skills only, and 60.19 for their unrestricted combination. The fixed expert skill reaches 86.67, below the three dynamically updated configurations; continued task-derived updates therefore add value beyond an initial expert procedure.

\paragraph{From Feedback to a Reusable Procedure.}
The Claude Code trace in Appendix~\ref{app:complete_example} illustrates how feedback becomes an operational procedure. Task~04 exposed six issues: a gross-margin error, an omitted profit-quality ratio, incomplete margin attribution, missing receivables and inventory checks, and qualitative rather than numerical ROE attribution. Reflection translated them into nine edits spanning calculation validation, driver decomposition, asset-quality analysis, the DuPont procedure, failure conditions, and pre-delivery checks. For example, the revised skill specifies chain-substitution formulas for net margin, asset turnover, and the equity multiplier and retains a check that their contributions sum to the ROE change. The resulting procedure encodes financial variables, required evidence, and validation rules that apply across companies and reporting periods.

\paragraph{Diagnostic Feedback and Reference Answers.}
Holding all other settings fixed, Table~\ref{tab:diagnostic_comparisons} compares targeted diagnostic feedback with a complete reference answer. Deltas are diagnostic feedback minus reference answer, and $\Delta$Act.\ is the mean activation-count difference over a 120-task run.

\begin{table}[t]
\centering
\caption{Feedback design and scene isolation by agent scaffold.}
\label{tab:diagnostic_comparisons}
\label{tab:feedback}
\label{tab:cross_scene}
\small
\setlength{\tabcolsep}{4.0pt}
\begin{tabular}{lrrrrr}
\toprule
& \multicolumn{3}{c}{Diagnostic feedback vs. reference answer} & \multicolumn{2}{c}{Isolated $-$ mixed} \\
\cmidrule(lr){2-4}\cmidrule(lr){5-6}
Agent scaffold & $\Delta$Score $\uparrow$ & $\Delta$Comp. $\downarrow$ & $\Delta$Act. $\uparrow$ & $\Delta$Score & $\Delta$Comp. \\
\midrule
Claude Code  & +6.38 & $-0.07$ & +4 & $+2.84$ & $+0.06$ \\
Codex        & +7.22 & $\mathbf{-0.14}$ & +6 & $+3.46$ & $+0.09$ \\
GenericAgent & +3.95 & $-0.10$ & \textbf{+9} & $+1.09$ & $+0.02$ \\
Letta        & \textbf{+7.93} & $-0.06$ & -- & $+4.17$ & $+0.12$ \\
\bottomrule
\end{tabular}
\end{table}

Diagnostic feedback raises scores by 3.95--7.93 points, reduces compliance issues by 0.06--0.14 per task, and adds 4--9 activations for scaffolds with selective retrieval (Table~\ref{tab:diagnostic_comparisons}). Letta's interface loads memory on every task, so its activation count is fixed. A reference answer presents one valid solution. Diagnostic feedback instead turns omitted evidence, incomplete analysis, and compliance failures in the current response into explicit update targets, producing more later retrieval and higher task performance.

\paragraph{Capability Dimensions.}
We map every rubric item to one of five dimensions: information and evidence use, analysis and calculation, conclusions and recommendations, report quality, and financial compliance. For a quality dimension $d$, let $\mathcal{J}_{i,d}$ contain the rubric items assigned to $d$ for task $i$, with awarded points $p_{a,k,i,j}$ and maximum points $p^{\max}_{i,j}$. Its normalized score rate is
\begin{align}
R_{a,k,i,d}
&=
\frac{\sum_{j\in\mathcal{J}_{i,d}}p_{a,k,i,j}}
{\sum_{j\in\mathcal{J}_{i,d}}p^{\max}_{i,j}}, \\
\Delta R_{a,d}
&=
\frac{100}{KN}\sum_{k=1}^{K}\sum_{i=1}^{N}
\left(R^{\evolve}_{a,k,i,d}-R^{\noevolve}_{a,k,i,d}\right).
\end{align}
For compliance, let $c_{a,k,i}$ be the number of triggered issues and $G_{s(i)}$ the total number of compliance issues defined for task $i$'s scene. We report
\begin{equation}
\Delta R^{\mathrm{comp}}_a
=
\frac{100}{KN}\sum_{k=1}^{K}\sum_{i=1}^{N}
\left(
\frac{c^{\noevolve}_{a,k,i}}{G_{s(i)}}
-
\frac{c^{\evolve}_{a,k,i}}{G_{s(i)}}
\right).
\end{equation}
Quality gain is the score rate with evolution minus the score rate without evolution. Compliance gain reverses this subtraction, so positive values indicate fewer normalized issue triggers. All five gains are reported in percentage points.

Every scaffold improves across all five dimensions (Figure~\ref{fig:evolution_diagnostics}, right). Report quality has the largest gain for each scaffold (6.85--18.17 percentage points), while normalized compliance gains range from 1.93 to 6.74 points. Report-level practices in structure, coverage, and expression accumulate across tasks more readily than case-specific evidence handling, analysis, and conclusions. GenericAgent records the smallest gains across dimensions, matching its lower overall gain and activation.

\paragraph{Cross-Scene Transfer and Interference.}
Interleaved scenes expose agents to both broad transfer and retrieval competition \citep{wei2025evomemory,wang2026evomembench}. We derive scene-isolated runs from all three streams by extracting the six tasks from each of the 20 scenes in their original relative order and executing them from a fresh agent state. The comparison holds the tasks and within-scene order constant and varies whether experience from intervening scenes is retained. Using 360 outputs per scaffold and condition, we report $\Delta\mathrm{Score}=S_{\mathrm{isolated}}-S_{\mathrm{mixed}}$ and $\Delta\mathrm{Comp.}=C_{\mathrm{isolated}}-C_{\mathrm{mixed}}$, where $S$ and $C$ denote mean score and compliance count; positive values favor isolation in score but mixed execution in compliance.

Across all four scaffolds, isolation raises scores by 1.09--4.17 points but triggers 0.02--0.12 more compliance issues per task (Table~\ref{tab:diagnostic_comparisons}). Interleaving therefore produces a consistent tradeoff: restricting experience to one scene improves task-specific quality, whereas retaining experience across scenes improves compliance. Interleaving carries general practices such as evidence checking and scope control across scenes, while isolation keeps retrieval focused on scene-specific procedures.

\section{Conclusion}

FinEvo-Bench evaluates whether agents improve through continued use on recurring professional work. Its 120 real-case financial tasks pair reusable workflows with widely varying inputs, analytical demands, and conclusions, using expert-authored rubrics to assess open-ended task quality and compliance. Across three interleaved streams, four scaffolds outperform state-reset controls by 9.33--19.37 points and 0.12--0.44 compliance issues per task. Diagnostic traces show how feedback becomes reusable procedures and how memory and skill configurations trade performance and cost. Full-benchmark scene isolation reveals a tradeoff between task-specific quality and cross-scene compliance transfer. Future extensions can apply this recurring-workflow design to additional backbones and professional domains beyond finance.

\FloatBarrier
\clearpage
\subsection*{AI use statement}

We used OpenAI Codex to assist with software code and experimental implementation, and to edit the manuscript for readability, \LaTeX{} formatting, and table and figure layout. All AI-assisted work was reviewed by the authors, and all AI-assisted code was tested by the authors before use. We did not use generative AI to generate synthetic datasets; clean or reformat data; develop theoretical models or conceptual frameworks; formulate mathematical claims or proofs; propose or refine hypotheses; translate research materials; conduct qualitative or thematic analysis; interpret results; or design or provide feedback on the research methodology or experiments. The remaining required disclosure categories are not applicable to this work. The authors take responsibility for the final content of the paper, including all text, claims, code, and artifacts produced with the assistance of generative AI.

\bibliographystyle{iclr2027_conference}
\bibliography{references}

\clearpage
\appendix
\raggedbottom
\captionsetup{hypcap=false}
\section{Benchmark Construction: A Worked Example of Financial Statement Analysis}
\label{app:construction_case}

This appendix documents the construction process from Section~3 for the \emph{Financial Statement Analysis} scene and Task~3. It traces the scene sources and design of six tasks to the input files, reference answer, scoring rubric, and review process.

\subsection{Scene Sources, Specification, and Construction of Six Tasks}
\label{app:scene_and_tasks}

\paragraph{Construction sources.}
Each scene $s$ uses the three sources defined in Section~3: a scene description $D_s$ provided by an institution, a reference professional procedure $P_s$ validated in practice, and a candidate pool $\mathcal{C}_s$ drawn from cases supplied by institutions or described in public sources. Table~\ref{tab:app_source_roles} lists their roles in task construction.

\begin{center}
\begin{minipage}{\linewidth}
\centering
\captionof{table}{Construction sources and their roles in the worked scene.}
\label{tab:app_source_roles}
{
\small
\setlength{\tabcolsep}{3.2pt}
\renewcommand{\arraystretch}{1.08}
\begin{tabularx}{\columnwidth}{@{}>{\raggedright\arraybackslash}p{0.20\columnwidth} >{\raggedright\arraybackslash}X >{\raggedright\arraybackslash}X@{}}
\toprule
Source & Information extracted & Role in construction \\
\midrule
Scene description $D_s$ & Business setting, professional roles, objective, and scope & Defines the business problem addressed by the scene \\
Reference procedure $P_s$ & Required input files, professional operations and checks, expected output, and professional constraints & Specifies how the work is performed and reviewed \\
Candidate cases $\mathcal{C}_s$ & Concrete facts, cross-file relations, risk patterns, judgment conditions, and possible conclusions & Provide the facts and data used to construct the six tasks \\
\bottomrule
\end{tabularx}
}

\end{minipage}
\end{center}

We review the source permissions and release conditions of each candidate case before admitting it to the eligible pool $\mathcal{C}^{\mathrm{eligible}}_s$.

\paragraph{Scene specification.}
A domain expert consolidates the scene description, reference procedure, and eligible cases into the specification with five components defined in Section~3. Table~\ref{tab:app_scene_spec} gives the result for Financial Statement Analysis.

\begin{center}
\begin{minipage}{\linewidth}
\centering
\captionof{table}{Extracted specification for the Financial Statement Analysis scene.}
\label{tab:app_scene_spec}
{
\small
\setlength{\tabcolsep}{3.3pt}
\renewcommand{\arraystretch}{1.08}
\begin{tabularx}{\columnwidth}{@{}>{\raggedright\arraybackslash}p{0.11\columnwidth} >{\raggedright\arraybackslash}p{0.22\columnwidth} >{\raggedright\arraybackslash}X@{}}
\toprule
Element & Meaning & Instantiation in the worked scene \\
\midrule
$O_s$ & Objective and scope & Assess corporate financial health for credit due diligence, post-loan review, large-credit assessment, or risk warning, without replacing the final credit decision \\
$E_s$ & Required input files & Financial statements, statement notes, audit reports, company information, and industry comparisons provided with the task \\
$A_s$ & Operations and checks & Verify data quality; calculate and interpret financial indicators; analyze profitability, assets, liabilities, cash flow, and DuPont drivers; scan warning signals; reconcile statements; form a supported conclusion \\
$Y_s$ & Expected deliverable & A structured financial-health report with calculations, findings supported by source data, a risk rating, and credit or monitoring recommendations \\
$K_s$ & Constraints & Use only supplied data; mark unavailable information; distinguish facts from inference; avoid unsupported assurances or guarantees \\
\bottomrule
\end{tabularx}
}

\end{minipage}
\end{center}

The reference procedure determines the work covered by the tasks and scoring rubric. Closely related steps may be combined, and optional operations apply only when the task requires them. Table~\ref{tab:app_procedure_mapping} maps the procedure to scoring requirements.

\begin{center}
\begin{minipage}{\textwidth}
\centering
\captionof{table}{Reference procedure and corresponding scoring requirements.}
\label{tab:app_procedure_mapping}
{
\small
\setlength{\tabcolsep}{3.6pt}
\renewcommand{\arraystretch}{1.08}
\begin{tabularx}{\textwidth}{@{}>{\raggedright\arraybackslash}p{0.13\textwidth} >{\raggedright\arraybackslash}X >{\raggedright\arraybackslash}X >{\raggedright\arraybackslash}p{0.14\textwidth}@{}}
\toprule
Procedure stage & Professional requirement & Task and scoring requirement & Rubric coverage \\
\midrule
Input and quality review & Confirm the entity, period, accounting basis, available statements and notes, audit opinion, and material changes in policy or consolidation scope. & Identify the input files provided, mark unavailable information, and report any non-standard audit opinion. & Data quality: 8 \\
Profitability and earnings quality & Calculate profitability ratios and distinguish accounting profit from cash realization and recurring operating performance. & Calculate and interpret the required ratios, explain changes, and compare them with the supplied industry data. & Profitability: 15 \\
Asset and liability quality & Examine receivables, inventory, special assets, debt structure, liquidity, and maturity matching. & Support findings with the relevant statements and notes and explain their business implications. & Asset quality: 12 \\
Cash-flow analysis & Analyze operating, investing, and financing cash flows; calculate free cash flow; interpret their joint pattern. & Cover all three cash-flow categories, operating-cash-flow quality, free cash flow, and the combined pattern. & Cash flow: 13 \\
DuPont decomposition & Decompose ROE and attribute changes to margin, turnover, and leverage. & Assess the decomposition, change attribution, and sustainability of the main driver. & DuPont: 10 \\
Warning review & Scan reporting warning signals and check scene-specific risk conditions. & Check the defined warning signals, possible reporting problems, and the eight risk conditions. & Warning review: 15 \\
Valuation when applicable & Add relative or absolute valuation only when the business request requires it. & Apply valuation criteria only to tasks that require valuation. & Not universally scored \\
Cross-statement reconciliation & Verify consistency across the balance sheet, income statement, cash-flow statement, and notes. & Check the defined relations and explain any unresolved discrepancy. & Reconciliation: 5 \\
Synthesis and delivery & Produce a health assessment, prioritized risks, quantitative support, monitoring actions, and a professional report. & Assess the conclusion, quantitative support, professional boundaries, and report quality. & Conclusions: 12; report: 10 \\
\bottomrule
\end{tabularx}
}

\end{minipage}
\end{center}

\paragraph{Construction of six tasks.}
Experts select six cases from the eligible pool based on the scene's business characteristics and professional procedure. The cases differ in company condition, input files, analytical focus, judgment conditions, or conclusions; cases with only superficial differences are excluded. Table~\ref{tab:app_task_family} lists the six Financial Statement Analysis cases.

\begin{center}
\begin{minipage}{\textwidth}
\centering
\captionof{table}{The six cases used for Financial Statement Analysis. File counts refer to the input files released with each task.}
\label{tab:app_task_family}
{
\small
\setlength{\tabcolsep}{3.1pt}
\renewcommand{\arraystretch}{1.08}
\begin{tabularx}{\textwidth}{@{}c >{\raggedright\arraybackslash}p{0.18\textwidth} >{\raggedright\arraybackslash}p{0.20\textwidth} >{\raggedright\arraybackslash}X c >{\raggedright\arraybackslash}p{0.11\textwidth}@{}}
\toprule
Task & Business setting & Input files & Main analytical focus & Files & Difficulty \\
\midrule
1 & Healthy mature manufacturer seeking a working-capital facility & Complete three-year statements, audit report, notes, and industry benchmark & Full-process baseline: profitability, solvency, cash flow, DuPont analysis, reconciliation, and credit recommendation & 10 & Easy \\
2 & Cyclical manufacturer under post-loan review & Complete statements with selected notes and sector benchmark & Separate firm-specific deterioration from an industry downturn; assess efficiency and cash-flow pressure & 9 & Medium \\
3 & Fast-growing materials company requesting a larger credit line & Complete statements, audit report, revenue and receivable notes, related-party information, non-recurring items, and benchmark & Assess earnings quality, collection risk, leverage-driven ROE, and credit capacity despite headline growth & 10 & Medium--hard \\
4 & Listed manufacturer under financial-reporting scrutiny & Complete statements plus cash, construction-in-progress, other-receivable, and related-party notes & Detect multiple warning signals, verify cross-statement consistency, and assess the level of risk & 10 & Hard \\
5 & Trading company with recent delinquency and severe deterioration & Complete statements, non-standard audit report, borrowing, receivable, contingent-liability, and benchmark files & Assess losses, leverage, debt service, negative operating cash flow, and multiple risk conditions & 9 & Hard \\
6 & Private manufacturer with severely incomplete reporting & One-year income statement and closing balance sheet plus company context & Limit the analysis to the available input files, avoid imputation, identify unavailable analyses, and request follow-up files & 3 & Easy--medium \\
\bottomrule
\end{tabularx}
}

\end{minipage}
\end{center}

Each selected case becomes a task with a natural-language request, input files, task metadata, and a reviewed reference answer. All six tasks follow the same professional procedure, but their facts and conclusions come from their own input files.

\subsection{Complete Worked Task: Task 3}
\label{app:task3}

Task~3 concerns Ruiheng New Materials and its request for a CNY~120 million credit line. The agent receives the following request:

\begin{quote}
\textbf{User request.} ``Ruiheng New Materials is applying for a CNY~120 million credit line. It has been growing quickly: revenue has increased for three consecutive years and profit is also rising. However, its cash flow appears problematic, and operating cash flow was negative last year. Please analyze its financial condition and assess whether the available financial information supports the requested credit amount.''
\end{quote}

Table~\ref{tab:app_task3_visibility} lists the components of Task~3 and their visibility.

\begin{center}
\begin{minipage}{\linewidth}
\centering
\captionof{table}{Task~3 components and their visibility to the agent.}
\label{tab:app_task3_visibility}
{
\small
\setlength{\tabcolsep}{3.2pt}
\renewcommand{\arraystretch}{1.08}
\begin{tabularx}{\columnwidth}{@{}>{\raggedright\arraybackslash}p{0.25\columnwidth} >{\raggedright\arraybackslash}X >{\raggedright\arraybackslash}p{0.22\columnwidth}@{}}
\toprule
Component & Contents and purpose & Visible to agent \\
\midrule
User request & Business question, requested amount, and initial concern & Yes \\
Input files & Ten files containing company information, financial statements, notes, an audit report, and comparison data & Yes \\
Task metadata & Variant type, difficulty, and dataset identifiers & No \\
Reference answer & One reviewed analysis and conclusion & No \\
Scoring rubric & Criteria, score bands, and professional and compliance checks used by the judge & No \\
\bottomrule
\end{tabularx}
}

\end{minipage}
\end{center}

The task includes ten input files. Table~\ref{tab:app_task3_trace} maps each file group to the required analysis and scoring criteria.

\begin{center}
\begin{minipage}{\textwidth}
\centering
\captionof{table}{Task~3 input-file-to-criterion mapping.}
\label{tab:app_task3_trace}
{
\small
\setlength{\tabcolsep}{3.5pt}
\renewcommand{\arraystretch}{1.08}
\begin{tabularx}{\textwidth}{@{}>{\raggedright\arraybackslash}p{0.14\textwidth} >{\raggedright\arraybackslash}p{0.25\textwidth} >{\raggedright\arraybackslash}X >{\raggedright\arraybackslash}p{0.19\textwidth}@{}}
\toprule
Input-file group & Files & Relevant information and required analysis & Rubric links \\
\midrule
Business context & \texttt{company\_\allowbreak{}overview.\allowbreak{}md} & Provides the requested amount and purpose, banking relationship, growth history, and management explanation. & P9-1--P9-4 \\
Core statements & \texttt{balance\_\allowbreak{}sheet.\allowbreak{}md}; \texttt{income\_\allowbreak{}statement.\allowbreak{}md}; \texttt{cashflow\_\allowbreak{}statement.\allowbreak{}md} & Revenue rises by 25\% while operating cash flow falls to CNY~$-80$ million and net receivables rise by 75\%; the report must calculate, reconcile, and interpret the divergence. & P2-1--P2-3; P3-1; P4-1--P4-4; P5-1--P5-3; P8-1--P8-2 \\
Audit report & \texttt{audit\_\allowbreak{}report.\allowbreak{}md} & Provides the audit opinion and the emphasis on receivable collection, which must be considered together with the statements and notes. & P1-2--P1-3; P3-1; P6-1 \\
Statement notes & \texttt{notes\_\allowbreak{}revenue.\allowbreak{}md}; \texttt{notes\_\allowbreak{}receivables.\allowbreak{}md}; \texttt{notes\_\allowbreak{}related\_\allowbreak{}party.\allowbreak{}md}; \texttt{notes\_\allowbreak{}non\_\allowbreak{}recurring.\allowbreak{}md} & Receivables older than one year reach 34.07\%, and non-recurring gains equal 38.18\% of net profit; the analysis must consider customer concentration, related-party transactions, allowances, and recurring profit. & P2-2--P2-3; P3-1; P6-1--P6-3 \\
Comparison data & \texttt{industry\_\allowbreak{}benchmark.\allowbreak{}md} & Industry profitability, earnings-quality, solvency, efficiency, and growth values provide the comparison basis for judging whether headline performance is sustainable. & P2-4; P3-4; P5-3; P9-3 \\
\bottomrule
\end{tabularx}
}

\end{minipage}
\end{center}

A complete response identifies available and unavailable input files; analyzes profitability, earnings quality, assets, liabilities, cash flow, and DuPont drivers; reviews financial warning signals; reconciles the statements; and gives a quantitatively supported rating and credit recommendation within the stated professional scope.

The reviewed reference answer rates the company as \emph{Concern}. It recommends against the full CNY~120 million request and proposes a CNY~60 million one-year facility with collateral, staged disbursement, and monitoring conditions. The cited reasons are negative operating cash flow, rapidly growing and aging receivables, the material contribution of non-recurring gains to profit, and leverage-driven ROE. Other recommendations receive credit if they are quantitatively supported, address the material risks, and remain within professional scope.

\subsection{Scene-Level Rubric and Quality Control}
\label{app:rubric_qc}

\paragraph{Rubric derivation.}
The six tasks share a 100-point rubric derived from the professional procedure. It scores correct use of the input files, completion of the necessary analysis, and evidence-supported conclusions across valid structures and formulations.

\begin{center}
\begin{minipage}{\textwidth}
\centering
\captionof{table}{Rubric allocation for the Financial Statement Analysis scene.}
\label{tab:app_rubric_allocation}
{
\small
\setlength{\tabcolsep}{3.5pt}
\renewcommand{\arraystretch}{1.06}
\begin{tabularx}{\columnwidth}{@{}>{\raggedright\arraybackslash}X r@{}}
\toprule
Criterion group & Points \\
\midrule
Data confirmation and quality assessment & 8 \\
Profitability and earnings quality & 15 \\
Asset quality and liability structure & 12 \\
Cash-flow analysis & 13 \\
DuPont analysis & 10 \\
Financial warning and reporting-risk review & 15 \\
Cross-statement reconciliation & 5 \\
Integrated assessment and conclusions & 12 \\
Report quality & 10 \\
\midrule
Total & 100 \\
\bottomrule
\end{tabularx}
}

\end{minipage}
\end{center}

Table~\ref{tab:app_scoring_examples} gives representative scoring rules. Applicability depends on the task input files and business request. A required analysis cannot be marked non-applicable when the necessary information is available. Numerical checks use the source data and defined formula, allowing a rounding tolerance below 0.5 percentage points or a clearly stated valid alternative convention.

\begin{center}
\begin{minipage}{\linewidth}
\centering
\captionof{table}{Representative scoring rules for individual criteria.}
\label{tab:app_scoring_examples}
{
\small
\setlength{\tabcolsep}{3.8pt}
\renewcommand{\arraystretch}{1.08}
\begin{tabularx}{\textwidth}{@{}>{\raggedright\arraybackslash}p{0.15\textwidth} >{\raggedright\arraybackslash}X >{\raggedright\arraybackslash}X >{\raggedright\arraybackslash}p{0.21\textwidth}@{}}
\toprule
Criterion & Full credit & Partial or zero credit & Non-applicable rule \\
\midrule
P2-2 Earnings quality (5) & The deliverable calculates cash collection, profit-to-cash conversion, and recurring-profit share and interprets them using the defined thresholds. & Two measures: 3; one measure or measures without interpretation: 1; no earnings-quality analysis: 0. & Applicable when the required statement values are present. \\
P3-1 Receivables (4) & The deliverable analyzes turnover or aging, allowance adequacy, and customer concentration using the available notes. & Material but incomplete analysis: 2; no receivables analysis: 0. & Full credit only when the package provides no receivables detail and the deliverable correctly states that the analysis is unavailable. \\
P5-2 Change attribution (4) & The deliverable attributes changes in ROE using chain substitution or an equivalent complete method. & Abbreviated attribution: 2; no attribution: 0. & Full credit when only one period is supplied and change attribution is impossible. \\
P9-4 Decision support (3) & The deliverable gives an actionable credit or monitoring recommendation supported by the task input files and avoids unsupported assurance. & Generic recommendation: 2; no recommendation: 0. & Full credit only when the business request genuinely does not call for a decision recommendation. \\
\bottomrule
\end{tabularx}
}

\end{minipage}
\end{center}

The rubric also checks for fabricated or unsupported information, conclusions that exceed the available input files, decisions beyond the agent's professional role, and unsupported assurances about credit approval or investment returns.

\begin{center}
\begin{minipage}{\textwidth}
\centering
\captionof{table}{Professional and financial-compliance checks in the scene rubric.}
\label{tab:app_compliance}
{
\small
\setlength{\tabcolsep}{3pt}
\renewcommand{\arraystretch}{1.0}
\begin{tabularx}{\textwidth}{@{}l >{\raggedright\arraybackslash}p{0.20\textwidth} >{\raggedright\arraybackslash}p{0.16\textwidth} >{\raggedright\arraybackslash}X@{}}
\toprule
ID & Check & Type & Observable condition in the final deliverable \\
\midrule
RED-1 & Fabricated financial data & Data use and compliance & Introduces a financial value, ratio, or audit opinion absent from the task input files. \\
RED-2 & Omitted core analysis & Critical task quality & Omits at least two of profitability analysis, cash-flow analysis, and warning review. \\
RED-3 & Skipped or missed risk checks & Critical professional quality & Omits the R1--R8 review or misses a condition explicitly supported by the task input files. \\
RED-4 & Severe-risk misclassification & Critical professional quality & Gives a favorable rating despite clear high-risk information and provides no corresponding warning. \\
RED-5 & Unsupported imputation & Data use and compliance & Fills missing core financial data by unsupported estimation instead of marking the data unavailable. \\
RED-6 & Missing scope statement & Professional scope & Does not state that the report supports analysis but does not constitute a final credit approval decision. \\
RED-7 & Prohibited assurance & Financial compliance & Guarantees approval or returns, or issues an unsupported directive to invest or lend. \\
\bottomrule
\end{tabularx}
}

\end{minipage}
\end{center}

\paragraph{Quality control.}
Domain experts first review the scene and task settings, the completeness of the input files, and the correctness of the reference answers. One expert then drafts the rubric from the scene specification and professional procedure. A second applies it to all six tasks, checking case applicability, coverage of required business steps, and consistency with the input files and reference answers. The experts correct any omission or conflict and review the affected material again. Table~\ref{tab:app_quality_control} lists these checks.

\begin{center}
\begin{minipage}{\textwidth}
\centering
\captionof{table}{Review steps for task construction and scoring.}
\label{tab:app_quality_control}
{
\small
\setlength{\tabcolsep}{3.8pt}
\renewcommand{\arraystretch}{1.08}
\begin{tabularx}{\textwidth}{@{}>{\raggedright\arraybackslash}p{0.15\textwidth} >{\raggedright\arraybackslash}p{0.23\textwidth} >{\raggedright\arraybackslash}X >{\raggedright\arraybackslash}p{0.18\textwidth}@{}}
\toprule
Stage & Review target & Checks & Action when revision is needed \\
\midrule
Source screening & Candidate cases & Source permissions and release conditions & Exclude cases that do not meet the release conditions \\
Scene definition & $(O_s,E_s,A_s,Y_s,K_s)$ & Consistency with the scene description, professional procedure, and cases & Revise the scene specification before selecting tasks \\
Task construction & Request, input files, metadata, and reference answer & Reasonable task setting, complete task package, correct calculations and conclusions, and substantive variation across cases & Correct the task and repeat the review \\
Rubric drafting & Criteria and grading rules & Coverage of the required business steps, clear score bands, and appropriate non-applicable conditions & Revise the criterion or grading rule \\
Rubric review & Rubric applied to all six tasks & Applicability to each case and consistency with the input files and reference answers & Revise the affected task or rubric and review it again \\
\bottomrule
\end{tabularx}
}

\end{minipage}
\end{center}

At evaluation time, the judge receives the task input files, the agent's final output, and the scoring rubric. Each criterion is applied to the submitted output using the task evidence.

\paragraph{Complete criterion inventory.}
Tables~\ref{tab:app_rubric_i} and~\ref{tab:app_rubric_ii} list all 32 scored criteria: 27 task criteria and five report-quality criteria. The P7 prefix is omitted because valuation is optional and does not apply to all six tasks.

\begin{center}
\begin{minipage}{\textwidth}
\centering
\captionof{table}{Complete scene-level criterion inventory, Part I: data quality through DuPont analysis (58 points).}
\label{tab:app_rubric_i}
{
\small
\setlength{\tabcolsep}{3pt}
\renewcommand{\arraystretch}{0.96}
\begin{tabularx}{\textwidth}{@{}l >{\raggedright\arraybackslash}p{0.20\textwidth} >{\raggedright\arraybackslash}X r@{}}
\toprule
ID & Criterion & Full-credit condition & Pts. \\
\midrule
P1-1 & Data completeness & The deliverable identifies the available statements, notes, and audit report and marks missing items without unsupported imputation. & 3 \\
P1-2 & Audit opinion & The deliverable correctly identifies the audit opinion and prominently reports any non-standard opinion; if no audit report is supplied, it states that the opinion is unverified. & 3 \\
P1-3 & Policy and scope changes & The deliverable checks material accounting-policy and consolidation-scope changes and explains their effect when the relevant information is available. & 2 \\
\midrule
P2-1 & Core profitability metrics & Correctly calculate gross margin, core operating margin, net margin, ROE, and other defined core indicators from the supplied data. & 4 \\
P2-2 & Earnings quality & Calculate and interpret cash collection, profit-to-cash conversion, and recurring-profit share using the defined thresholds. & 5 \\
P2-3 & Profitability drivers & Attribute margin changes and analyze material expense-ratio movements rather than merely listing values. & 3 \\
P2-4 & Industry comparison & Compare the core profitability indicators with the supplied industry benchmark or explicitly state that comparison data are unavailable. & 3 \\
\midrule
P3-1 & Receivables & Analyze turnover or aging, allowance adequacy, and customer concentration using the available notes; flag an over-one-year share above 30\%. & 4 \\
P3-2 & Inventory & Analyze inventory turnover, impairment allowance, and consistency with revenue using the available information. & 3 \\
P3-3 & Goodwill and special assets & Evaluate goodwill or long-running construction in progress and associated impairment or capitalization risk when applicable. & 2 \\
P3-4 & Debt structure and solvency & Analyze interest-bearing debt, maturity structure, liquidity ratios, and interest coverage. & 3 \\
\midrule
P4-1 & Three cash-flow categories & Analyze operating, investing, and financing cash flow and identify the material movements in each. & 4 \\
P4-2 & Operating-cash-flow quality & Reconcile operating cash flow with reported profit and connect the result to earnings-quality measures. & 4 \\
P4-3 & Free cash flow & Correctly calculate free cash flow and interpret its trend. & 2 \\
P4-4 & Joint cash-flow pattern & Identify and explain the combined operating, investing, and financing cash-flow pattern and its credit implications. & 3 \\
\midrule
P5-1 & Three-factor DuPont & Correctly decompose ROE into net margin, asset turnover, and equity multiplier. & 4 \\
P5-2 & Change attribution & Use chain substitution or an equivalent method to attribute changes in ROE. & 4 \\
P5-3 & Driver sustainability & Identify the dominant ROE driver and assess whether it is sustainable. & 2 \\
\bottomrule
\end{tabularx}
}

\end{minipage}
\end{center}

\begin{center}
\begin{minipage}{\textwidth}
\centering
\captionof{table}{Complete scene-level criterion inventory, Part II: warning review, reconciliation, synthesis, and report quality (42 points).}
\label{tab:app_rubric_ii}
{
\small
\setlength{\tabcolsep}{3pt}
\renewcommand{\arraystretch}{0.96}
\begin{tabularx}{\textwidth}{@{}l >{\raggedright\arraybackslash}p{0.20\textwidth} >{\raggedright\arraybackslash}X r@{}}
\toprule
ID & Criterion & Full-credit condition & Pts. \\
\midrule
P6-1 & Warning-signal scan & The deliverable reports checks of at least six defined warning types and assigns a severity supported by the task input files. & 5 \\
P6-2 & Reporting-risk review & The deliverable checks income, cost, profit, and cash-flow information for signs of aggressive reporting or manipulation. & 5 \\
P6-3 & Risk conditions R1--R8 & The deliverable reports checks of all eight defined risk conditions and surfaces any triggered condition in the report summary. & 5 \\
\midrule
P8-1 & Cross-statement reconciliation & Verify the defined relations among cash, profit, equity movements, and statement totals. & 3 \\
P8-2 & Discrepancy handling & Confirm consistency or identify, localize, and qualify any unresolved discrepancy. & 2 \\
\midrule
P9-1 & Financial-health rating & Give a clear rating whose severity is consistent with the preceding analysis. & 3 \\
P9-2 & Strengths and risks & Summarize concrete, case-specific strengths and risks rather than generic observations. & 3 \\
P9-3 & Quantitative support & Support qualitative conclusions with values, changes, periods, and relevant comparisons. & 3 \\
P9-4 & Decision support & Give an actionable credit or monitoring recommendation supported by the task input files and avoid unsupported assurance. & 3 \\
\midrule
FMT-1 & Clear presentation & Present the main findings and numerical results clearly; tables may be used when appropriate. & 2 \\
FMT-2 & Analytical coverage & Cover at least seven of the eight required analytical components. Section titles and ordering may vary. & 3 \\
FMT-3 & Period and source labels & Mark reporting periods and sources for central data and conclusions. & 2 \\
FMT-4 & Scope statement & State that the report is for decision support and does not constitute a final credit approval decision. & 1 \\
FMT-5 & Data fidelity & Use only values supported by the task input files. & 2 \\
\bottomrule
\end{tabularx}
}

\end{minipage}
\end{center}

\section{Three-Stage Evaluation Pipeline and Prompt Templates}
\label{app:pipeline_prompts}

This appendix documents the three-stage harness used in the longitudinal evaluation. Stage~1 executes a task, Stage~2 scores the resulting deliverable, and Stage~3 supplies targeted diagnostic feedback to the evaluated agent for reflection. Stages~1 and~3 run one of the four evaluated agent frameworks (Claude Code, Codex, Letta, or GenericAgent), whereas Stage~2 always uses the same Claude Code judge. For each task, Stage~3 resumes the exact conversation created in Stage~1.

\subsection{Pipeline Overview}

Table~\ref{tab:app_pipeline_roles} summarizes the executor, workspace, and output of each stage. The task instruction and user request are concatenated into the Stage~1 message sent to the agent. Stage~2 writes the specific reasons for lost credit to \texttt{judge.md}, which is then added to the Stage~1 workspace for reflection.

\begin{center}
\begin{minipage}{\textwidth}
\centering
\captionof{table}{Roles, workspace visibility, and outputs in the three-stage evaluation pipeline.}
\label{tab:app_pipeline_roles}
{
\small
\setlength{\tabcolsep}{4pt}
\renewcommand{\arraystretch}{1.12}
\begin{tabularx}{\textwidth}{@{}>{\raggedright\arraybackslash}p{0.09\textwidth} >{\raggedright\arraybackslash}p{0.17\textwidth} >{\raggedright\arraybackslash}p{0.22\textwidth} >{\raggedright\arraybackslash}X >{\raggedright\arraybackslash}p{0.17\textwidth}@{}}
\toprule
Stage & Function & Executor & Visible workspace & Primary output \\
\midrule
1 & Task execution & One of four evaluated agent frameworks & \texttt{input/}, containing only the files supplied with the task & \texttt{model\_\allowbreak{}result.\allowbreak{}md} \\
2 & Rubric scoring & Fixed Claude Code judge & \texttt{input/}, \texttt{RUBRIC.\allowbreak{}md}, and \texttt{model\_\allowbreak{}result.\allowbreak{}md} & \texttt{judge.\allowbreak{}md} \\
3 & Feedback reflection & The same agent framework in the exact Stage~1 conversation & The Stage~1 workspace, with \texttt{judge.\allowbreak{}md} added & Updated agent state \\
\bottomrule
\end{tabularx}
}

\end{minipage}
\end{center}

\begingroup
\setlength{\fboxsep}{4pt}
\noindent\fbox{%
\begin{minipage}{0.965\textwidth}
\textbf{Algorithm B.1: Three-stage longitudinal evaluation}
\vspace{1pt}

\small
\begin{tabularx}{\linewidth}{@{}r >{\raggedright\arraybackslash}X@{}}
\textbf{Input:} & ordered task stream $(\tau_1,\ldots,\tau_N)$; one of the four evaluated frameworks $a$; task rubrics; fixed Claude Code judge $J$ \\
\textbf{State:} & agent state $S_a$ accumulated across the task stream \\
\midrule
1 & \textbf{for} each task $\tau_i$ in stream order \textbf{do} \\
2 & \quad Create \texttt{input/}; concatenate the task instruction and user request as message $p_i$. \\
3 & \quad Start conversation $C_i$ with $a$, send $p_i$, and save the Markdown deliverable as \texttt{model\_\allowbreak{}result.\allowbreak{}md}. \\
4 & \quad In an isolated workspace containing \texttt{input/}, \texttt{RUBRIC.\allowbreak{}md}, and \texttt{model\_\allowbreak{}result.\allowbreak{}md}, invoke $J$ and save the specific reasons for lost credit as \texttt{judge.\allowbreak{}md}. \\
5 & \quad Add \texttt{judge.\allowbreak{}md} to the Stage~1 workspace. \\
6 & \quad Resume the exact conversation $C_i$, send the Stage~3 prompt, and allow $a$ to update $S_a$. \\
7 & \textbf{end for} \\
\end{tabularx}
\end{minipage}}
\endgroup

\subsection{Stage 1: Task Execution}

\begingroup
\noindent\begin{minipage}{\linewidth}
\setlength{\fboxsep}{4pt}
\noindent\fbox{%
\begin{minipage}{0.965\textwidth}
\colorbox{black}{\parbox{0.965\linewidth}{\color{white}\bfseries Stage 1 Prompt Panel: Execute the Current Task}}
\vspace{3pt}

\small
\noindent\textbf{Prompt template}
\vspace{3pt}

\noindent\begin{minipage}{\linewidth}
\small\ttfamily\raggedright
Complete the task described below using the files in input/.
Write the final deliverable as model\_result.md in the current directory.
After the file has been written, reply only ``done.''
\par\vspace{5pt}
[Task instruction]
\par\vspace{3pt}
[User request]
\end{minipage}
\end{minipage}}
\captionof{figure}{Stage~1 task-execution prompt.}
\label{fig:app_stage1_prompt}
\end{minipage}
\endgroup

\subsection{Stage 2: Fixed Claude Code Scoring}

\begingroup
\noindent\begin{minipage}{\linewidth}
\setlength{\fboxsep}{4pt}
\noindent\fbox{%
\begin{minipage}{0.965\textwidth}
\colorbox{black}{\parbox{0.965\linewidth}{\color{white}\bfseries Stage 2 Prompt Panel: Score with the Fixed Claude Code Judge}}
\vspace{3pt}

\small
\noindent\textbf{Prompt template}
\vspace{3pt}

\noindent\begin{minipage}{\linewidth}
\small\ttfamily\raggedright
You are a strict evaluator. The current workspace contains the original task
inputs under input/, the scoring rubric in RUBRIC.md, and the agent deliverable
in model\_result.md.
\par\vspace{4pt}
Evaluate model\_result.md strictly according to RUBRIC.md. Identify every
deficiency that reduces the score and support it with evidence from the
deliverable and, when needed, the task inputs.
\par\vspace{4pt}
Write concise diagnostic feedback to judge.md, describing the specific reasons
for lost credit. After the file has been written,
reply only ``done.''
\end{minipage}
\end{minipage}}
\captionof{figure}{Stage~2 scoring prompt.}
\label{fig:app_stage2_prompt}
\end{minipage}
\endgroup

\subsection{Stage 3: Feedback Reflection in the Same Conversation}

\begingroup
\noindent\begin{minipage}{\linewidth}
\setlength{\fboxsep}{4pt}
\noindent\fbox{%
\begin{minipage}{0.965\textwidth}
\colorbox{black}{\parbox{0.965\linewidth}{\color{white}\bfseries Stage 3 Prompt Panel: Reflect on Feedback}}
\vspace{3pt}

\small
\noindent\textbf{Prompt template}
\vspace{3pt}

\noindent\begin{minipage}{\linewidth}
\small\ttfamily\raggedright
The file judge.md contains diagnostic feedback on your previous response.
Read judge.md and reflect on the task; you may update your state and
accumulate experience based on it.
After you have finished, reply only ``done.''
\end{minipage}
\end{minipage}}
\captionof{figure}{Stage~3 reflection prompt.}
\label{fig:app_stage3_prompt}
\end{minipage}
\endgroup

\section{Complete Example of Feedback-Driven Experience Consolidation}
\label{app:complete_example}

This appendix follows one Task~04 trace from the Financial Statement Analysis scene. The task asked the evaluated Claude Code agent to prepare a risk-oriented analysis of an anonymized biomedical company's 2021--2023 consolidated statements, audit reports, notes, and industry reference data. Stage~2 wrote six rubric-grounded reasons for lost credit to \texttt{judge.md}. Stage~3 received this file for reflection, producing six memory-change blocks and nine skill-change blocks. The trace follows the three-stage protocol in Appendix~\ref{app:pipeline_prompts}.

\subsection{Stage 1 Deliverable}

The agent produced an eight-section financial-analysis report covering data quality, profitability, asset quality, cash flow, DuPont analysis, financial warnings, statement reconciliation, and an overall credit conclusion. The excerpts below are translated from the archived output and show the content involved in the subsequent feedback.

\begingroup
\setlength{\fboxsep}{4pt}
\noindent\fbox{%
\begin{minipage}{0.965\textwidth}
\textbf{Selected evidence from the Stage~1 deliverable}
\vspace{5pt}

\small
\begin{tabularx}{\linewidth}{@{}>{\bfseries}>{\raggedright\arraybackslash}p{0.19\linewidth} >{\raggedright\arraybackslash}X@{}}
Profitability table & ``Gross margin: 42.00\% (2023), 40.02\% (2022), and 40.00\% (2021).'' \\
Profit quality & The report explicitly calculated the cash collection ratio and net cash ratio, but did not calculate core profit divided by total profit. \\
Receivables & The report analyzed turnover days and receivables growth against revenue growth, then noted possible year-end collection arrangements. \\
Inventory & The report compared turnover days with the industry value and discussed possible understatement or incomplete accounting. \\
ROE attribution & The report verified the three-factor product and stated that leverage was the main driver, but gave no numerical contribution for each factor. \\
\end{tabularx}
\end{minipage}}
\endgroup

\subsection{Stage 2 Evaluation and Diagnostic Feedback}

Table~\ref{tab:app_c_feedback} presents the six diagnostic findings supplied for Stage~3 reflection.

\begin{center}
\begin{minipage}{\textwidth}
\centering
\captionof{table}{Diagnostic feedback supplied for Task~04.}
\label{tab:app_c_feedback}
{\small
\setlength{\tabcolsep}{4.5pt}
\renewcommand{\arraystretch}{1.13}
\begin{tabularx}{\textwidth}{@{}>{\raggedright\arraybackslash}p{0.035\textwidth} >{\raggedright\arraybackslash}p{0.20\textwidth} >{\raggedright\arraybackslash}X@{}}
\toprule
No. & Topic & Diagnostic feedback \\
\midrule
1 & Core profitability metric & The 2022 gross margin was reported as 40.02\%; the correct value is 41.00\%, beyond the 0.5-percentage-point tolerance. \\
2 & Profit-quality assessment & Only the cash collection ratio and net cash ratio were explicitly calculated. The report omitted core profit divided by total profit and its interpretation. \\
3 & Profitability-driver decomposition & The report listed changes in gross margin and expense ratios but did not systematically attribute them, for example through a quantity--price decomposition. \\
4 & Receivables analysis & The report omitted allowance sufficiency and customer concentration, and did not mark the missing receivables-aging note as unavailable. \\
5 & Inventory analysis & The report omitted inventory-impairment analysis and the relation between inventory and revenue, and did not mark the missing inventory detail as unavailable. \\
6 & Change attribution & The report verified the three-factor ROE product and gave a directional interpretation, but did not use chain substitution or an equivalent method to quantify each factor's contribution. \\
\bottomrule
\end{tabularx}}

\end{minipage}
\end{center}

\subsection{Stage 3 Memory Update}

Stage~3 resumed the conversation used to generate the report. The memory update added six items: exact gross-margin recalculation, complete coverage of three profit-quality ratios, systematic gross-margin attribution, receivables allowance and concentration checks, inventory impairment and inventory--revenue matching, and numerical ROE-factor attribution. Missing case details are recorded as unavailable within the corresponding checks.

\subsection{Stage 3 Skill Update}

The skill update translated the same feedback into nine executable additions spanning calculation validation, driver decomposition, asset-quality checks, the DuPont procedure, failure catalogues, and final quality checks. Table~\ref{tab:app_c_skill_update} shows the operational differences before and after the update.

\begin{center}
\begin{minipage}{\textwidth}
\centering
\captionof{table}{Operational changes made to the reusable financial-analysis skill after Task~04.}
\label{tab:app_c_skill_update}
{\small
\setlength{\tabcolsep}{4.5pt}
\renewcommand{\arraystretch}{1.14}
\begin{tabularx}{\textwidth}{@{}>{\raggedright\arraybackslash}p{0.20\textwidth} >{\raggedright\arraybackslash}X >{\raggedright\arraybackslash}X@{}}
\toprule
Procedure block & Before Task~04 & Added after Task~04 \\
\midrule
Calculation validation & Validate core-profit arithmetic and reconcile the result with operating profit. & Recalculate gross margin as $(\mathrm{revenue}-\mathrm{cost})/\mathrm{revenue}$ and explicitly compute the cash collection ratio, net cash ratio, and core profit/total profit. \\
Gross-margin attribution & Discuss raw-material prices, product mix, pricing power, competition, and cost pass-through. & Require a systematic decomposition, including quantity and price contributions when supported, rather than a list of directional changes. \\
Receivables & Compare receivables growth, revenue growth, aging, allowance practice, and related-party collection periods. & Require allowance-sufficiency and top-customer-concentration analyses; mark unavailable aging-note data explicitly. \\
Inventory & Compare growth and turnover with revenue and industry values; include inventory-impairment analysis. & Require an explicit inventory--revenue matching test, impairment sufficiency, and an unavailable-data label when inventory details are absent. \\
ROE attribution & Check that the three-factor product equals ROE and that the contributions sum to the total change. & Require numerical chain-substitution contributions for net margin, asset turnover, and the equity multiplier; a product check and directional conclusion alone are insufficient. \\
\bottomrule
\end{tabularx}}

\end{minipage}
\end{center}

The skill also adds the six Task~04 failure modes to its section-level catalogues and incorporates them into its consolidated pre-delivery checklist.

\subsection{Representative Persisted Procedure}

The updated DuPont block provides a concrete example of an executable rule created during reflection. Let $m_t$, $u_t$, and $e_t$ denote net margin, asset turnover, and the equity multiplier in period $t$. For a base period~0 and current period~1, the skill contains the following procedure:

\begingroup
\noindent\begin{minipage}{\linewidth}
\setlength{\fboxsep}{4pt}
\noindent\fbox{%
\begin{minipage}{0.965\textwidth}
\colorbox{black}{\parbox{0.965\linewidth}{\color{white}\bfseries Representative Skill Excerpt: Chain-Substitution Procedure}}
\vspace{4pt}

\small
\begin{align*}
\Delta_{\mathrm{margin}} &= (m_1-m_0)u_0e_0, \\
\Delta_{\mathrm{turnover}} &= m_1(u_1-u_0)e_0, \\
\Delta_{\mathrm{leverage}} &= m_1u_1(e_1-e_0), \\
\Delta\mathrm{ROE} &= \Delta_{\mathrm{margin}}+\Delta_{\mathrm{turnover}}+\Delta_{\mathrm{leverage}}.
\end{align*}

\noindent\textbf{Required validation.} Compute each contribution numerically, verify that their sum equals the observed ROE change, and recheck the inputs if the identity does not close. Report both the contribution values and their business interpretation.
\end{minipage}}
\captionof{figure}{Chain-substitution instructions added to the updated skill after Task~04.}
\label{fig:app_c_final_skill_roe}
\end{minipage}
\endgroup

\end{document}